\documentclass[10pt,twocolumn,letterpaper]{article}

\usepackage[pagenumbers]{cvpr} 

\usepackage{multirow}
\usepackage{tabularx}
\usepackage{url}

\newcolumntype{V}{>{\raggedright\arraybackslash}X}
\newcolumntype{Y}{>{\centering\arraybackslash}X}
\newcolumntype{Z}{>{\raggedleft\arraybackslash}X}

\newcolumntype{v}{>{\hsize=.4\hsize}V}
\newcolumntype{y}{>{\hsize=.4\hsize}Y}
\newcolumntype{z}{>{\hsize=.4\hsize}Z}

\newcolumntype{l}{>{\hsize=.3\hsize}V}
\newcolumntype{m}{>{\hsize=.3\hsize}Y}
\newcolumntype{n}{>{\hsize=.3\hsize}Z}

\definecolor{cvprblue}{rgb}{0.21,0.49,0.74}
\usepackage[pagebackref,breaklinks,colorlinks,allcolors=cvprblue]{hyperref}

\usepackage[capitalize]{cleveref}
\crefname{section}{Sec.}{Secs.}
\Crefname{section}{Section}{Sections}
\Crefname{table}{Table}{Tables}
\crefname{table}{Tab.}{Tabs.}

\def\paperID{7} 
\def\confName{CVPR}
\def\confYear{2026}

\title{How to Choose Your Teacher for Fine Grained Image Recognition}

\author{
    \textbf{Oswin Gosal\textsuperscript{\dag}\thanks{Equal contribution.}, Edwin Arkel Rios\textsuperscript{\ddag}\footnotemark[1], Augusto Christian Surya\textsuperscript{\dag}, Fernando Mikael\textsuperscript{\dag},} \\
    \textbf{Bo-Cheng Lai\textsuperscript{\ddag}, Min-Chun Hu\textsuperscript{\dag}} \\
    \textsuperscript{\dag}\textit{National Tsing Hua University, Taiwan}, \textsuperscript{\ddag}\textit{National Yang Ming Chiao Tung University, Taiwan}, \\
}

\begin{document}
\maketitle

\begin{abstract}
\label{abstract}

\vspace{-0.7cm}

Fine-grained image recognition classifies subcategories such as bird species or car models. While state-of-the-art (SOTA) models are accurate, they are often too resource-intensive for deployment on constrained devices. Knowledge distillation addresses this by transferring knowledge from a large teacher model to a smaller student model. A key challenge is selecting the right teacher, as it heavily impacts student performance. This paper introduces a teacher selection metric, \textbf{Ratio 1-2}, based on teacher prediction ratios. Extensive analysis of over one thousand experiments across 3 students, 8 teachers, and 8 datasets under 4 training strategies demonstrates that our metric improves teacher selection by 18\% over previous methods, enabling small student models to achieve up to 17\% accuracy gains. Experiment codebase is available at: \href{https://github.com/arkel23/FGIR-KD-Teacher}{https://github.com/arkel23/FGIR-KD-Teacher}.

\end{abstract}

\section{Introduction}
\label{intro}
Fine-grained image recognition (FGIR) aims to classify similar subcategories within a super-category, such as bird species \cite{wah_caltech-ucsd_nodate} or car models \cite{krause_3d_2013}. While most work focus on specialized discriminative feature learning modules \cite{fu_look_2017, hu_see_2019, rao_counterfactual_2021, he_transfg_2022, rios_global-local_2025}, recent work by Ye et al.~\cite{ye_image_2024} shows that backbone architecture plays a critical role in recognition performance. In general, larger backbones achieve higher accuracy but require significant computational resources for training and deployment, limiting their applicability in real-world, resource-constrained environments \cite{violos_towards_2024, wang_cost-aware_2019}.

Knowledge distillation (KD) provides a practical solution by transferring knowledge from a large teacher model to a smaller student model \cite{hinton_distilling_2015}. However, the resulting student accuracy $Acc$ depends on multiple factors, including the dataset $D$, student architecture $S$, teacher $T$, training strategy and loss function $L$, and hyperparameters $H$. Prior work has studied the effects of training strategies and distillation losses \cite{romero_fitnets_2015, park_relational_2019, tian_contrastive_2019, violos_towards_2024}, but teacher selection remains a critical and underexplored challenge \cite{cho_efficacy_2019, tan_improving_2024}. Cho and Hariharan \cite{cho_efficacy_2019} demonstrated that higher teacher accuracy does not lead to better student performance, highlighting the need for better teacher evaluation metrics. Tan et al.~\cite{tan_improving_2024} proposed selecting teachers based on the \textbf{standard deviation of secondary soft probabilities}, where larger deviations indicate more informative soft targets. However, as shown in Section~\ref{sec_results}, this metric is less effective in FGIR scenarios, where inter-class differences are inherently subtle and secondary probability variations are naturally small.

Since teacher choice can significantly affect student performance, with up to 17\% accuracy difference observed on the Aircraft dataset, we propose a novel teacher selection metric, \textbf{Ratio 1-2}. As differences in FGIR are naturally subtle, \textbf{Ratio 1-2} measures teacher confidence by computing the ratio between the top two raw (non-normalized) logits, serving as an indicator of overconfidence. A more effective teacher produces more nuanced predictions, enabling the student to learn subtle inter-class distinctions rather than simply mimicking hard labels.

The contributions of this work are as follows:
\begin{itemize}[topsep=3pt, itemsep=3pt, parsep=3pt]
  \item We conduct an extensive study of over a thousand experiments across 8 datasets, 3 student and 8 teacher models under 4 training strategies, providing a comprehensive analysis of teacher impact in knowledge distillation.
  
  \item We introduce a novel metric, \textbf{Ratio 1-2}, based on the ratio of the top two raw teacher logits as a measure of prediction overconfidence. The metric shows consistently strong correlation with student performance in over 50\% of experimental settings and improves student accuracy by up to 9.4\% compared to existing metrics.
\end{itemize}

\section{Experiment Methodology}
\label{sec_methodology}

In this paper, we conduct a large-scale study to analyze the factors that impact the effectiveness of KD in fine-grained image recognition, in terms of student accuracy $Acc$:

\vspace{-0.3cm}

\begin{equation}
    Acc = f(D, T, S, L, H)
\end{equation}

where $f$ is a function of the target dataset $D$, the teacher $T$ , the student $S$, the training strategy and loss function $L$, and hyperparameters $H$. Since exploring all possible combinations of these factors is computationally unfeasible, and prior work has extensively studied the effect of $L$, we reduce the design space by primarily using the vanilla knowledge distillation loss proposed by Hinton et al.~\cite{hinton_distilling_2015} and fixed hyperparameters $H$, unless otherwise stated. 

\noindent\textbf{Datasets.} To evaluate the effect of the target dataset $D$, we conduct experiments on 8 fine-grained classification datasets: Aircraft \cite{maji_fine-grained_2013}, Cars \cite{krause_3d_2013}, CUB \cite{wah_caltech-ucsd_nodate}, Dogs \cite{khosla_novel_2011}, Flowers \cite{nilsback_automated_2008}, Moe \cite{noauthor_tagged_nodate}, NABirds \cite{van_horn_building_2015}, and Pets \cite{parkhi_cats_2012}.

\noindent\textbf{Teachers.} Students are trained by 8 teachers, including convolutional- and attention-based models: VGG-19 \cite{simonyan_very_2014}, ResNet-101 (RN-101) \cite{he_deep_2016}, ResNetV2-101 (RNV2-101) \cite{he_identity_2016}, ResNetV2-101x3-BiT (RN-101x3) \cite{kolesnikov_big_2020}, ViT B-16 \cite{dosovitskiy_image_2020}, Swin-B \cite{liu_swin_2021}, ConvNeXt-Base (CNX-B) \cite{liu_convnet_2022}, and VAN-B3 \cite{guo_visual_2023}. To cover varying levels of fine-grained specialization, teachers are trained using three strategies: (a) frozen backbone with a linear classifier (\textbf{FZ}), (b) full fine-tuning (\textbf{FT}), and (c) counterfactual attention learning \textbf{(CAL)} \cite{rao_counterfactual_2021}, a method based on data-aware augmentations. 

\noindent\textbf{Students.} We evaluate three student architectures covering both convolutional and transformer-based models: ResNet-18 \cite{he_deep_2016}, LCNet-35 \cite{cui_pp-lcnet_2021}, and LeViT-128S \cite{graham_levit_2021}. For ResNet-18, we conduct experiments under two initialization settings: pretrained on ImageNet-1k and trained from scratch.

\noindent\textbf{Training Strategy and Loss Function.} We adopt the vanilla KD loss proposed by Hinton et al.~\cite{hinton_distilling_2015}:

\begin{equation}
    \label{eq_loss}
    \mathcal{L} = \mathcal{L}_{\text{CE}}(y^S, y_{gt}) + \beta \mathcal{L}_{\text{KD}}(y^S, y^T)
\end{equation}

where $y^S$ and $y^T$ denote the student and teacher predictions, $y_{gt}$ the ground-truth labels, $\mathcal{L}_{\text{CE}}$ the cross-entropy loss, and $\mathcal{L}_{\text{KD}}$ the KL-divergence loss.

Based on the teacher specialization, we group the student training strategies into four categories: a) students trained using frozen teachers \textbf{FZ}, b) students trained using fine-tuned teachers \textbf{FT}, c) students trained using CAL teachers \textbf{CAL}, and d) \textbf{TGDA:} a state-of-the-art FGIR distillation strategy, Teacher-Guided Data Augmentation \cite{rios_fine-grained_2025},  where CAL teachers generate data-aware augmentations to mitigate overfitting and improve student learning of subtle visual distinctions. All student models use the loss in Eq.~\ref{eq_loss}. For TGDA, an additional distillation term $\mathcal{L}_{\text{KD}}(y^S_{aug}, y^T_{aug})$ is applied to augmented images generated by the teacher.

\section{Indicators for Teacher Selection}
\label{sec_indicators}

\subsection{Previous metrics: Teacher Accuracy (TAC) and Secondary Soft Probabilities (SSP)}
\label{prev_metrics}

Assuming the target dataset $D$, student $S$, and training strategy $L$ are fixed, prior work \cite{cho_efficacy_2019, tan_improving_2024} has shown that the choice of teacher $T$ significantly affects the resulting student accuracy $Acc$. Therefore, identifying a reliable metric for evaluating teacher effectiveness is a key challenge in KD.

As a baseline, we consider \textbf{Teacher Accuracy (TAC)} \cite{cho_efficacy_2019}, defined as the classification accuracy of the teacher on the target dataset. TAC is simple and easy to compute; however, previous studies \cite{cho_efficacy_2019, tan_improving_2024} show that it is not a reliable predictor of student performance, as higher teacher accuracy does not necessarily translate to better outcomes.

To address this limitation, Tan et al.~\cite{tan_improving_2024} proposed the \textbf{Secondary Soft Probabilities (SSP)} metric, which measures the dispersion of secondary class probabilities to quantify how informative the teacher's soft predictions are. Given an input image $\mathbf{x}$ and a classifier $F$ with $N$ output classes, the logits are defined as $\mathbf{y} = F(\mathbf{x})$. Applying softmax and sorting the probabilities in descending order produces $\mathbf{Q} = \text{sort}(\text{Softmax}(\mathbf{y}))$ where $Q_1 \geq Q_2 \geq \cdots \geq Q_N$. The SSP metric is defined as the standard deviation of the second to fourth highest probabilities:

\begin{equation}
    \label{eq_ssp}
    SSP = \sqrt{\frac{\sum_{i=2}^{K+1}(Q_i - \mu)^2}{K}},
\end{equation}

where $\mu = \sum_{i=2}^{K+1} Q_i / K$ and $K=3$.

\subsection{Proposed Metric: Ratio 1-2}
\label{ssec_r12}

\begin{figure}[!htb]
\vspace{-0.3cm}
    \begin{center}
        \includegraphics[width=0.9\linewidth]{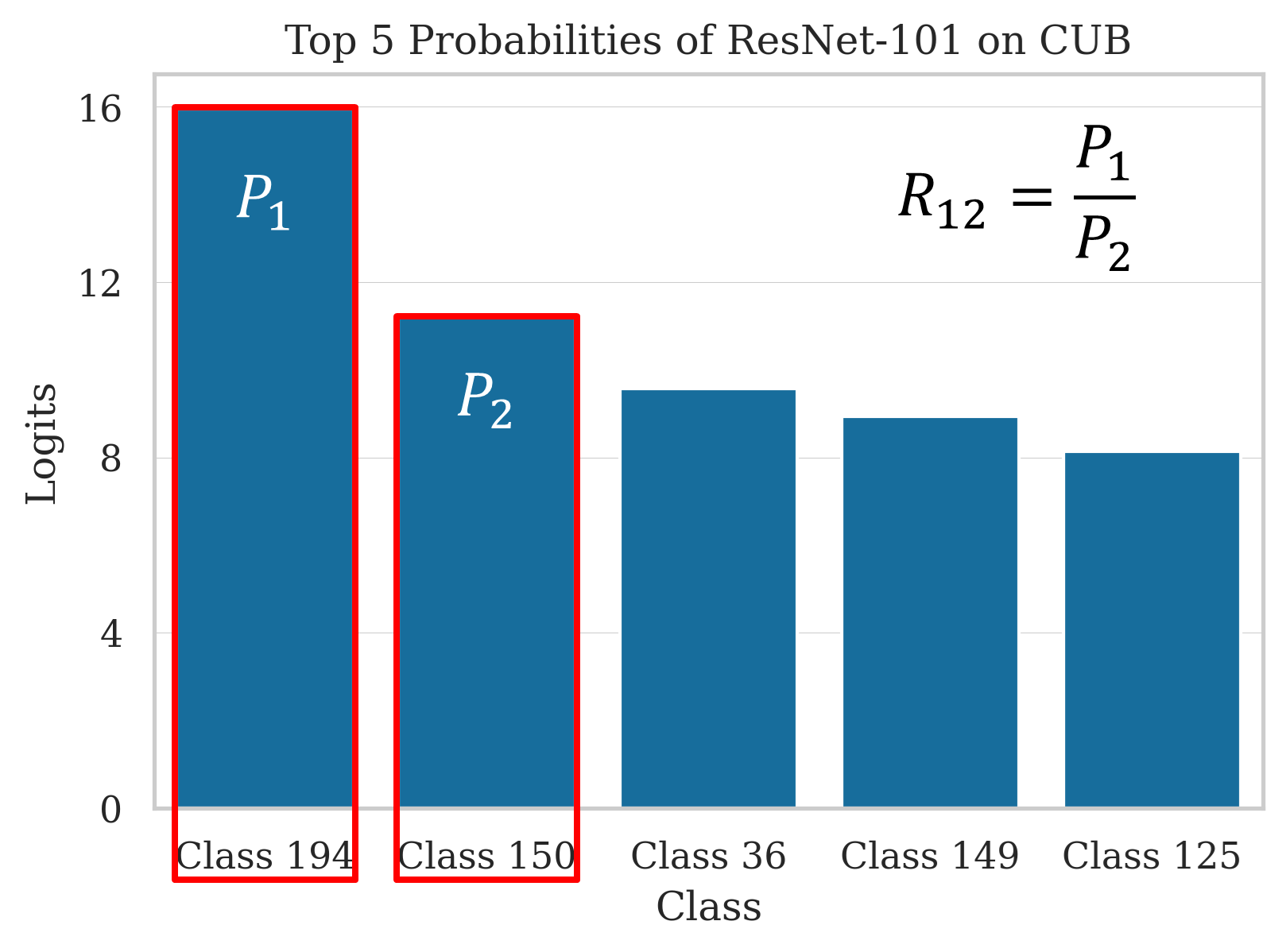}
    
    \end{center}
    \vspace{-0.3cm}
    \caption{Visualization of the top-1 and top-2 teacher  logits used in the $R_{12}$ metric. The figure shows the top-5 raw (non-normalized) logits from a ResNet-101 teacher on the CUB dataset.}
    \vspace{-0.3cm}
    \label{figure_r12}
\end{figure}

Motivated by the observation that FGIR involves subtle inter-class differences, we propose a novel metric based on the raw (non-normalized) logits of the teacher model. Specifically, we compute the ratio between the two highest logits in each prediction. These ratios reflect the teacher's confidence in its predictions. Intuitively, overconfidence can be detrimental, where small inter-class differences should be considered.

Given an input image $\mathbf{x}$ and a classifier $F$ with $N$ output classes, the logits are defined as $\mathbf{y} = F(\mathbf{x})$. The logits are sorted in descending order to obtain $\mathbf{P} = \text{sort}(\mathbf{y})$ such that $( P_1 \geq P_2 \geq \cdots \geq P_N $. The \textbf{Ratio 1-2} metric is defined as:

\begin{equation}
    \label{eq_r12}
    R_{12} = \frac{P_1}{P_2},
\end{equation}

\begin{table}[!htb]
    \centering
    \caption{Correlation groupings (\%) across all experiments.} 
    \vspace{-0.3cm}
    \begin{tabularx}{\linewidth}{Vzzz}
        \toprule
        Correlation & TAC \cite{cho_efficacy_2019} & SSP \cite{tan_improving_2024} & $R_{12}$ (Ours) \\
        \midrule
        Weak (0 – 0.5) & 42.2 & 39.8 & 28.1 \\
        Modest (0.51 – 0.7) & 25.8 & 16.4 & 21.1 \\
        Strong (0.71 – 1) & 32.0 & 43.8 & \textbf{50.8} \\
        \bottomrule
    \end{tabularx}    \label{table_corr_group_all_metrics}
\end{table}

\begin{table}[!htb]
    \centering
    \caption{Absolute average correlations across all student models, teacher models, and training strategies on all datasets.} 
    \vspace{-0.3cm}
    \begin{tabularx}{\linewidth}{VZZZ}
        \toprule
        Dataset & TAC \cite{cho_efficacy_2019} & SSP \cite{tan_improving_2024} & $R_{12}$ (Ours) \\
        \midrule
        Aircraft & 0.377 & 0.372 & \textbf{0.379} \\
        Cars & \textbf{0.479} & 0.447 & 0.407 \\
        CUB & 0.615 & 0.284 & \textbf{0.717} \\
        Dogs & 0.368 & \textbf{0.864} & 0.831 \\
        Flowers & 0.682 & 0.654 & \textbf{0.726} \\
        Moe & 0.529 & 0.626 & \textbf{0.628} \\
        NABirds & 0.570 & 0.469 & \textbf{0.626} \\
        Pets & 0.568 & \textbf{0.759} & 0.715 \\
        \midrule
        Average & 0.524 & 0.559 & \textbf{0.629} \\
        \bottomrule
    \end{tabularx}    \label{table_corr_avg_all_metrics}
    \vspace{-0.3cm}
\end{table}

where $P_1$ and $P_2$ denote the highest and second-highest logits, respectively. A large ratio indicates that the teacher strongly favors a single class, which may reduce the amount of information available to the student. In contrast, a smaller ratio suggests that the teacher considers multiple classes plausible, providing richer signals that better capture fine-grained class relationships. The final $R_{12}$ score for a teacher is computed by averaging the ratio across all samples within each batch and over all training epochs.

\section{Results and Discussion}
\label{sec_results}

In total, we conducted 1,216 experiments across 8 datasets using 3 student models, 8 teacher models, 4 training strategies, and 4 additional distillation losses. To quantify the effectiveness of each teacher-side metric, we computed the correlation between teacher evaluation metrics and the resulting student accuracies across all experimental settings. Correlation strengths were categorized into three levels: weak (0–0.50), modest (0.51–0.70), and strong (0.71–1.00), following the classification proposed by Hinkle et al.~\cite{hinkle_applied_2003}. To ensure robustness against outliers, particularly those introduced by suboptimal teacher models, we employed Spearman rank correlation, which is more resilient to non-linear relationships and outliers \cite{mukaka_guide_2012}.

\subsection{Overall Results and Trends}

\Cref{table_corr_group_all_metrics} summarizes the performance of the three evaluated teacher-side metrics across all training settings and datasets. We also computed the absolute average correlations for each metric and reported the results grouped by dataset in \Cref{table_corr_avg_all_metrics}. Our method achieves the highest correlation on 5 out of the 8 datasets, and attains the highest overall average correlation of 0.629. Moreover, it yields the largest proportion of strong correlations (50.8\%), representing a 7\% improvement over the second-best metric. These results further highlight the limitations of previous metrics, which do not account for subtle distinctions in FGIR.

\begin{figure}[!htb]
    \vspace{-0.3cm}
    \begin{center}
        \includegraphics[width=0.9\linewidth]{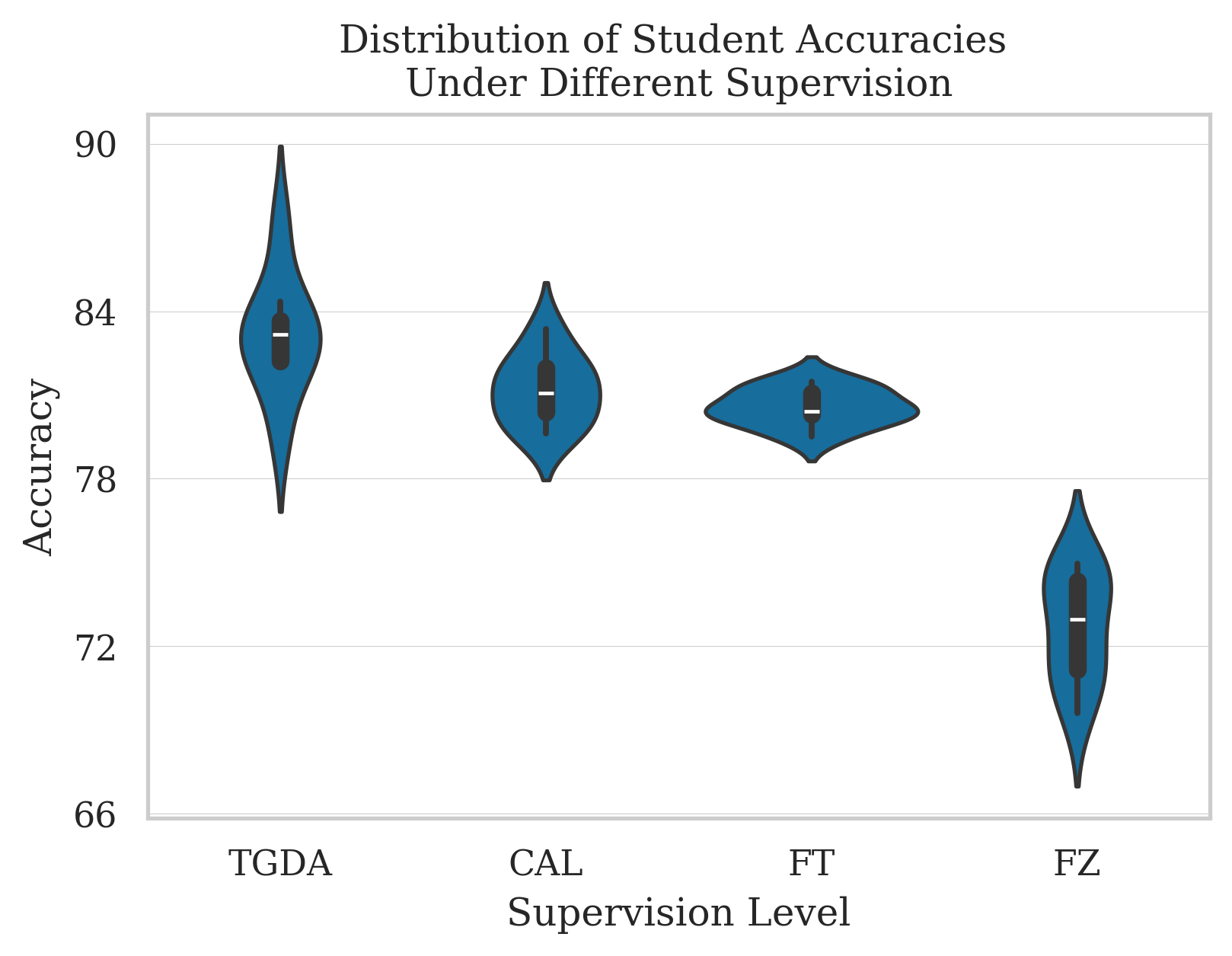}
    \end{center}
    \vspace{-0.3cm}
    \caption{Comparison of ResNet-18 accuracies trained from scratch on the Aircraft dataset.}
    \label{figure_supervision}
    \vspace{-0.3cm}
\end{figure}

\subsection{Effect of Training Strategy}

\begin{table}[!htb]
    \vspace{-0.3cm}
    \centering
    \caption{Correlation groupings (\%) across supervision levels.}
    \vspace{-0.3cm}
    \begin{tabularx}{\linewidth}{XZZZ}
        \toprule
        Correlation & TAC \cite{cho_efficacy_2019} & SSP \cite{tan_improving_2024} & $R_{12}$ \\
        \midrule
        {\textit{CAL}} &  &  &  \\
        Weak & 37.5 & 41.7 & 20.8 \\
        Modest & 33.3 & 20.8 & 20.8 \\
        Strong & 29.2 & 37.5 & \textbf{58.3} \\
        \midrule
        {\textit{TGDA}} &  &  &  \\
        Weak & 41.7 & 29.2 & 12.5 \\
        Modest & 41.7 & 20.8 & 20.8 \\
        Strong & 16.7 & 50.0 & \textbf{66.7} \\
        \bottomrule
    \end{tabularx}    \label{table_corr_group_all_supervision}
\end{table}

As shown in \Cref{figure_supervision}, we observe that selecting an appropriate teacher can, in some cases, allow lower-supervision settings to achieve performance levels comparable to those under higher supervision. These findings highlight the critical role of teacher selection in KD.

As shown in \Cref{table_corr_group_all_supervision}, in the CAL setting $\mathbf{R_{12}}$ outperforms the other metrics with the highest overall average correlation of 0.674 and the highest proportion of strong correlations at 58.3\%. A similar trend is observed in the TGDA setting, where the average correlation increases to 0.753, and the proportion of strong correlations rises to 66.7\%. \textit{These results suggest that the proposed metric is particularly effective for specialized teachers}.

\subsection{Teacher Impact}

\begin{figure}[!htb]
    \vspace{-0.3cm}
    \begin{center}
        \includegraphics[width=0.9\linewidth]{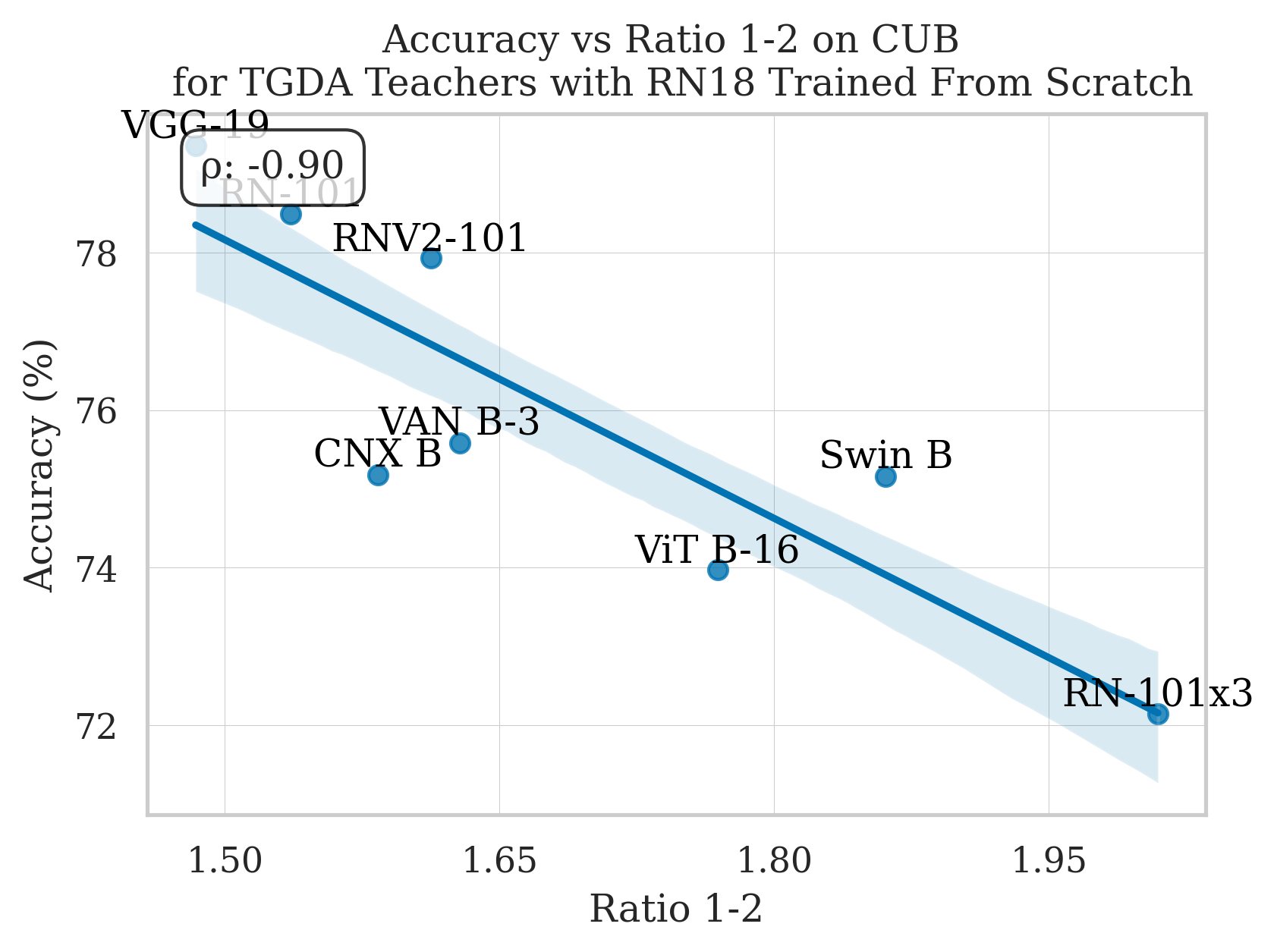}
    \end{center}
    \vspace{-0.3cm}
    \caption{Distribution of the teachers’ $R_{12}$ on the CUB under the TGDA setting, using a ResNet-18 trained from scratch.}
    \vspace{-0.3cm}    \label{figure_r12_tgda_rn18_np_cub}

\end{figure}

We further examine the behavior of individual teachers and their impact on student performance. \Cref{figure_r12_tgda_rn18_np_cub} presents the distribution of the proposed $\mathbf{R_{12}}$ for each teacher when training a ResNet-18 student under the TGDA setting on CUB. Among the teachers, VGG-19 exhibits the lowest average ratio, while ResNetV2-101x3-BiT the highest.

To generalize this observation, we analyze the teachers with the lowest $\mathbf{R_{12}}$ values across datasets. Notably, VGG-19 consistently appears among the lowest-ratio teachers on multiple datasets, being Aircraft, CUB, Dogs, Moe, and Pets, followed by ResNet-101 on Cars and Flowers, and ResNetV2-101 on NABirds. These models are among the smallest in our teacher pool, with VGG-19 having approximately 23 million parameters and both ResNet-101 and ResNetV2-101 having around 55 million parameters on CUB. This trend suggests that smaller models tend to produce less overconfident predictions, making them more effective as teachers in the distillation process.

Conversely, ResNetV2-101x3-BiT consistently yields the highest $\mathbf{R_{12}}$ values across datasets. This model, which is the largest in our teacher pool, has approximately 421 million parameters on CUB. Due to their increased capacity, large models are capable of extracting rich feature representations, often achieving high standalone accuracy. However, this capacity also leads to highly confident predictions, which may reduce the informativeness of the soft targets provided to students. This observation aligns with the findings of Cho and Hariharan \cite{cho_efficacy_2019}, who reported that students often fail to learn effectively from excessively large teachers due to the mismatch in capacity. \textit{These findings reinforce the idea that while large teachers are powerful, their overconfidence can hinder effective knowledge transfer, highlighting the importance of balancing model capacity with suitable confidence levels for optimal teacher selection.}

\subsection{Ratio 1-2 Enable Training Better Students}

\begin{table}[!htb]
    \vspace{-0.3cm}
    \centering
    \caption{Accuracy (\%) of a LCNet-35 student trained under TGDA using teachers selected by each metric.}
    \vspace{-0.3cm}
    \begin{tabularx}{\linewidth}{vnzzn}
        \toprule
        Dataset & CE & TAC \cite{cho_efficacy_2019} & SSP \cite{tan_improving_2024} & {$R_{12}$}\\
        \midrule
        Aircraft & 77.3 & 84.0 & 84.5 & \textbf{85.2} \\
        Cars & 29.9 & 75.7 & \textbf{82.5} & \textbf{82.5} \\
        CUB & 51.2 & 67.0 & 64.1 & \textbf{73.5} \\
        Dogs & 43.9 & 55.2 & \textbf{68.0} & \textbf{68.0} \\
        Flowers & 70.8 & 77.6 & \textbf{88.6} & \textbf{88.6} \\
        Moe & 90.6 & 92.4 & \textbf{95.2} & \textbf{95.2} \\
        NABirds & 22.9 & 62.4 & 62.4 & \textbf{67.8} \\
        Pets & 61.3 & 78.6 & 79.1 & \textbf{80.2} \\
        \midrule
        Average & 56.0 & 74.1 & 78.1 & \textbf{80.1} \\
        \bottomrule
    \end{tabularx}
    \vspace{-0.3cm}
    \label{table_acc_comp_ln35}
\end{table}

\Cref{table_acc_comp_ln35} reports the student accuracies obtained when an LCNet-35 student (trained from scratch) is paired with teachers selected by the baseline metrics and by our $\mathbf{R_{12}}$. \textbf{CE} denotes the baseline trained with cross-entropy loss without knowledge distillation. Across all datasets, students distilled from teachers selected using $\mathbf{R_{12}}$ consistently outperform both the CE baseline and those trained using teachers chosen by prior metrics. Notably, we observe substantial improvements of 52.5\% on Cars and 44.9\% on NABirds relative to CE. On average, the accuracy of the students reached 80.1\%, exceeding the averages achieved using teachers chosen by existing metrics by up to 6\%.

Despite the architectural mismatch between LCNet-35 and the diverse pool of teachers (which includes both CNN-based and transformer-based models), $\mathbf{R_{12}}$ reliably identifies the most suitable teacher for each dataset. \textit{These results demonstrate that the proposed $\mathbf{R_{12}}$ metric remains agnostic to discrepancies in model architectures}. Overall, these findings underscore the importance of correct teacher selection, particularly for small students, where choosing an appropriate teacher can yield drastic improvements.

\section{Conclusion}
\label{sec_conclusion}

In this work, we investigated knowledge distillation with a focus on teacher selection. We found that larger teachers, while more capable, are not always the most effective for teaching a student. We proposed a novel metric, $\mathbf{R_{12}}$, measuring teacher confidence via the ratio of top logits. Our results show that this metric successfully identifies suitable teachers in over 50\% of cases. Notably, we trained a much smaller student (LCNet-35) and improved its accuracy by up to 6\% on average compared to training with teachers selected by previous metrics, narrowing the gap between large and small models and making strides toward efficient yet accurate FGIR models.

{
    \bibliographystyle{ieeenat_fullname}
    \bibliography{references}
}


\end{document}